\documentclass[letterpaper, 10 pt, conference]{ieeeconf}  

\IEEEoverridecommandlockouts                              
\usepackage{graphics} 
\usepackage{epsfig} 
\usepackage{mathptmx} 
\usepackage{times} 
\usepackage{amsmath} 
\usepackage{amssymb}  
\usepackage{algpseudocode}
\usepackage{xcolor}
\usepackage{booktabs}
\usepackage{caption}

\newcommand{\E}{{\mathbb E}}
\newcommand{\R}{{\mathbb R}}

\newcommand{\until}[1]{\{1,\dots, #1\}}
\newcommand*\mc[0]{\mathcal}        

\newtheorem{rem}{Remark}
\newtheorem{alg}{Algorithm}
\title{\LARGE \bf
Classical vs. Bayesian methods for linear system identification: \\
point estimators and confidence sets}

\author{D. Romeres, G. Prando,  G. Pillonetto and A. Chiuso $^\dagger$
\thanks{This work has been partially supported by the FIRB project ``Learning meets time'' (RBFR12M3AC) funded by MIUR.}
\thanks{$^\dagger$ Dept. of Information  Engineering, University of Padova (e-mail: \{\tt \small romeresd,prandogi,giapi,chiuso\}@dei.unipd.it)}%
}

\begin{document}

\maketitle
\thispagestyle{empty}
\pagestyle{empty}

\begin{abstract}

This paper compares classical parametric methods with recently developed Bayesian methods for system identification. A \emph{Full Bayes}  solution is considered together with one of the standard approximations based on the \emph{Empirical Bayes}  paradigm. Results regarding point estimators for the impulse response as well as for confidence regions are reported. 

\end{abstract}


\section{Introduction} 
\label{sec: intro}
Linear system identification is sometimes considered to be a mature field, see e.g. \cite{Ljung,Soderstrom}. In particular  parametric prediction error methods (PEM) are by now well developed and understood.  Yet, facing in an effective manner the so-called bias variance dilemma trading model complexity vs. data fit is still an open issue and, very recently,  regularization methods for system identification \cite{DoanLSER1984,KitagawaTAC1985,Goodwin1992} have been revitalized; see e.g. \cite{SS2010,BVAR2010,SS2011,SurveyKBsysid}. 

In particular experimental evidence has shown that parametric methods may give rather unreliable results when model complexity is not fixed but has rather to be determined from data. Since most criteria for determining complexity are derived using asymptotic arguments, this is yet another symptom suggesting  that asymptotic theory is not to be blindly trusted. This is not only related to issues pertaining to local minima (as discussed for instance in \cite{Garatti:2004}), but also to the fact that it is difficult to say ``how much data is enough data'' to be in the asymptotic regime.  These issues concerning asymptotic results become even more dramatic when parameter estimation has to be coupled with model selection, resulting in so called Post Model Selection Estimators (PMSE). \cite{Leeb} have pointed out that asymptotic analysis is rather delicate in this case. 

Therefore, if under certain circumstances asymptotic analysis fails in delivering reliable indications as to the variability of an estimator, how would one go about providing, e.g., confidence sets for estimated systems? This is certainly of primary importance in a system identification exercise as one is not only interested in providing estimators for some quantity of interest, but also in providing quality tags which measure how reliable an estimator is. 

In this paper we shall compare Bayesian methods in a non-parametric setting to classical parametric approaches. 
In particular, we will compare the uncertainty sets which can be found following the classical parametric paradigm, specifically PEM equipped with BIC criterion and with an oracle to estimate model complexity, with the non-parametric approaches, both following the \emph{Full Bayes} as well as the \emph{Empirical Bayes} methods.

The paper is organized as follows: Section \ref{sec: Problem Formulation} introduces the system identification problem, while Sections \ref{sec: Parametric Identification Methods} and \ref{sec: Non-Parametric Identification Methods} respectively illustrate the classical parametric methods and the Bayesian non-parametric approach adopted in a system identification setting; both the point estimators and the confidence sets arising from these two approaches are presented. Section \ref{sec: Simulations} provides an experimental comparison of these techniques, while Section \ref{sec: Conclusions} draws some final remarks on the observed results.

\section{Problem Formulation}
\label{sec: Problem Formulation} 
Consider, for the sake of the exposition, a single-input-single-output Output Error model:
\begin{equation}
\label{eq: OE model}
y(t)= [h \ast u](t) + e(t)
\end{equation}
where $y(t), \, u(t) \in \R$ are respectively the measurable input and output, $e(t)$ is a zero mean Gaussian white noise uncorrelated to $u(t)$ and $h(t)$ is the impulse response of the model.\\
Given a finite set of input-output data points $\mathcal{D}=\{u(t), y(t)\}_{t \in \until{T} }$, system identification aims at estimating the impulse response $h(t)$. Moreover, one could also be interested in determining a (random) set  which is likely to include the unknown true $h(t)$: this range is generally referred to as confidence set. In this paper we will compare the classical and the Bayesian methods for system identification on both these two aspects of the problem.\\
In the remaining of the paper, we shall consider $\{u(t)\}$ and $\{y(t)\}$ as jointly stationary zero-mean stochastic processes and denote with $U,\, Y \in \R^T$ the vectors with entries $u(t), \, y(t),\ t=1,...T$, respectively.\\


\section{Classical Identification Methods}
\label{sec: Parametric Identification Methods}

\subsection{Point estimator}
Within the classical parametric identification framework, one assumes that the system to be identified belongs to a specific model class $\mathcal{M}$  (e.g. ARMAX, OE, Box-Jenkins, state-space, etc.), which is parametrized through a parameter $\theta \in \Theta$, i.e. $\mathcal{M}(\theta)$. The commonly used PEM (Prediction Error Method) determines the estimate of $\theta$ by minimizing the sum of squared prediction errors, i.e.:
\begin{equation}\label{eq:theta_est}
\hat{\theta}_{PEM} = \arg \min_{\theta\in \Theta} J(\theta) = \arg \min_{\theta\in \Theta} \frac{1}{T} \sum_{t=1}^T (y(t)-\hat{y}(t|\theta))^2
\end{equation}
where $\hat{y}(t|\theta)$ denotes the one-step ahead predictor of the chosen model class. Once $\hat{\theta}_{PEM}$ has been determined, one can then compute the corresponding impulse response estimate $\hat{h}_{\theta_{PEM}}(t)$.

Many interesting properties of these estimators are derived using asymptotic arguments, i.e. considering $T\rightarrow\infty$. For instance, for Gaussian innovations $e(t)$ and for fixed model complexity, these methods have proved to be asymptotically efficient. However, model complexity, which strongly affects their effectiveness, has to be estimated from the data. Different approaches are commonly exploited for this purpose, such as Cross-Validation or the Information Criteria (AIC/FPE, BIC/MDL, etc.) which are derived by asymptotic arguments. From these considerations a natural question arises: how many data have to be considered for these asymptotic properties to be reliable in a finite-sample domain? The answer is not general and could be really application-dependent.

\subsection{Confidence Set}\label{subsec:pem_confidence_set}
\subsubsection{Asymptotic} 
\label{subsubsec:asymp_pem_confset}
Consider the estimate \eqref{eq:theta_est}; under the assumption that the true system belongs to the chosen model class $\mathcal{M}$ and some other mild assumptions, (e.g. $\hat{\theta}_{PEM}$ gives rise to a uniformly stable model and the given data $\left\{y(t)\right\},\left\{u(t)\right\}$ are jointly quasi-stationary signals), it holds that
\begin{equation}\label{eq:asdistr}
\hat{\theta}_{PEM} \rightarrow \mathcal{N}\left(\theta_0, \frac{\Sigma_\theta}{T}\right), \quad \mbox{as } T\rightarrow\infty
\end{equation}
where $\theta_0$ is the unique value in $\Theta$ such that
\begin{equation}
\hat\theta_{PEM} \rightarrow \theta_0,\quad \mbox{w.p. 1 as } T\rightarrow \infty 
\end{equation}
and
\begin{eqnarray}\label{equ:pem_as_cov}
\Sigma_\theta &=& \sigma^2 \left\{\lim_{T\rightarrow\infty} \frac{1}{T}\sum_{t=1}^T   \E\left[\psi(t,\theta_0)\psi(t,\theta_0)^\top \right] \right\}^{-1}\\
\psi(t,\theta_0) &=& \frac{d}{d\theta}\hat{y}(t|\theta)\vert_{\theta=\theta_0}
\end{eqnarray}
See \cite{Ljung:99} for more details.
\\Once $\hat{\theta}_{PEM}$ has been determined exploiting the given $T$ input-output pairs, the asymptotic covariance \eqref{equ:pem_as_cov} can be approximated as
\begin{eqnarray}\label{equ:approx_pem_as_cov}
\hat\Sigma_\theta \hspace*{-3mm}&=&\hspace*{-3mm} J(\hat\theta_T) \left\{\frac{1}{T}\sum_{t=1}^T   \E\left[\psi(t,\hat\theta_{PEM})\psi(t,\hat\theta_{PEM})^\top \right] \right\}^{-1}\\
\psi(t,\hat\theta_{PEM}) \hspace*{-3mm}&=&\hspace*{-3mm} \frac{d}{d\theta}\hat{y}(t|\theta)\vert_{\theta=\hat\theta_{PEM}}
\end{eqnarray}
Notice that, in case of Gaussian innovations $\Sigma_\theta$ coincides with the Cramer-Rao lower bound, thus proving the aforementioned asymptotic efficiency of the PEM estimators.

Observe that the asymptotic covariance \eqref{equ:approx_pem_as_cov} describes the (asymptotic) confidence set in the space of the estimated parameters $\theta$. For further comparison with the Bayesian methods, we are also interested in determining a confidence set for the estimated impulse response coefficients $\hat h_{\theta_{PEM}}$. To do this, one could proceed analytically by linearizing the map:
\begin{eqnarray}
\mathcal{L}: \Theta &\rightarrow& \mathbb{R}^n \\
\theta &\mapsto& h\nonumber
\end{eqnarray}
and thus directly mapping the parameter confidence set onto the space of impulse response coefficients. Notice that, for simplicity, we consider a truncated impulse response, where the length $n$ can be chosen in order to account only for the relevant part of the impulse response.

In order to avoid the linear approximation introduced by the mentioned approach, we prefer to resort to Monte-Carlo sampling which yields a  point  distribution of the confidence set in the impulse response space. We first draw $N$ samples $\theta^{(i)}$ from the distribution $p_T(\cdot)\sim \mathcal{N}\left(\hat\theta_{PEM},\frac{\hat{\Sigma}_\theta}{T}\right)$; for each of them we build the model $\mathcal{M}(\theta^{(i)})$ and we compute its impulse response $h_{\theta^{(i)}}$ of length $n$. We then determine the confidence set composed by the $h_{\theta^{(i)}}$ associated with the $\alpha$-fraction of the highest probability $p_T(\cdot)$, i.e:
\begin{equation}
\label{def: S1 PEM alpha}
S^{PEM+ASYMP}_{\alpha} = \left\{h_{\theta^{(i)}}: p_T(\theta^{(i)}) \geq p^{PEM+ASYMP}_{\alpha}, \theta^{(i)}\in\Theta \right\}
\end{equation}
where $p^{PEM+ASYMP}_{\alpha}$ is the $(1-\alpha)$-percentile of the set $\left\{p_T(\theta^{(i)})\right\},$\ ${i=[1,N]}$. 

\subsubsection{Likelihood Sampling  } As an alternative, instead of relying on the approximation \eqref{equ:approx_pem_as_cov} to the asymptotic covariance \eqref{equ:pem_as_cov}, one could define a confidence set sampling from the likelihood function $p(Y\vert \theta,\hat{\sigma}^2)$, with $\hat\sigma^2$ being a noise variance estimate (obtained e.g. through a Least-Squares model). 
%
%
In fact, assuming a flat prior distribution $p(\theta)$ for the parameters, the likelihood function is proportional to the posterior distribution:
\begin{equation}\label{equ:posterior_param}
p(\theta|Y,\hat\sigma^2) \propto p(Y|\theta,\hat\sigma^2) = (2\pi\hat\sigma^2)^{-T/2}\exp\left\{ -\frac{T}{2\hat\sigma^2}J(\theta)\right\}
\end{equation}
where $J(\theta)$ has been defined in \eqref{eq:theta_est}. Hence, we design an MCMC algorithm to obtain $N$ samples $\theta^{(i)}$ from \eqref{equ:posterior_param}. From these we compute the corresponding impulse responses $h_{\theta^{(i)}}$ and we define the set
\begin{equation}
\label{def: S PEM alpha}
S^{PEM+LIK}_{\alpha} = \left\{h_{\theta^{(i)}}: p(\theta^{(i)}|Y,\hat{\sigma}^2) \geq p^{PEM+LIK}_{\alpha},  \theta^{(i)}\in\Theta \right\}
\end{equation} 
where $p^{PEM+LIK}_{\alpha}$ is the $(1-\alpha)$-percentile of the set $\left\{p(\theta^{(i)}|Y,\hat\sigma^2)\right\},$\ ${i=[1,N]}$. 

As previously said, sampling techniques allow to avoid approximations of asymptotic expressions. However, they are still approximations of the true uncertainty associated to the estimated parameter $\hat\theta_{PEM}$. Indeed, for the definition of the previous confidence sets, it has been assumed that the model class $\mathcal{M}$ and the model complexity are fixed, even if in practice model selection is performed using the available data. That is, $\hat\theta_{PEM}$ is a so-called post-model-selection estimator (PMSE): in order to define a more accurate confidence set, we should take into account also the uncertainty related to the model selection step. However, as emphasized in \cite{Leeb}, the finite-sample distribution of a PMSE generally has a quite intricate shape; moreover, even if one tries to estimate it through a sampling method, one has to recall that the finite-sample distribution of a PMSE is not uniformly close to its asymptotic limit \eqref{equ:pem_as_cov}.


\section{Bayesian Identification Methods}
\label{sec: Non-Parametric Identification Methods}
\subsection{Point estimator}
\label{subsec: Bayesian Imp Resp Est}
Non-parametric approaches to the system identification problem follow the Bayesian framework: one postulates that the impulse response to be estimated, $h(t)$, is itself a random process and one seeks for its posterior distribution given the data, $p(h|Y)$. 

The a priori probability distribution given to $h(t)$ is called prior 
\begin{equation}
h \thicksim p(h\vert \eta)
\end{equation}
and in general depends upon some unknown parameters $\eta$, called hyperparameters hereafter, which need to be estimated from data.\\
A common and convenient choice is to model $h(t)$ as a zero mean Gaussian process, independent of the noise $e(t)$ with covariance function $K(t,s)$, i.e.
\begin{align*}
\E h(t) &= 0\\
\E h(t) h(s) &= K_\eta(t,s)
\end{align*}
The covariance function $K_\eta(t,s)$ is sometimes called kernel in the Machine Learning community. This type of Gaussian priors can be derived following Maximum Entropy arguments, see e.g. \cite{NicFer:1998:IFA_1779,CarliCL2015}.

The minimum variance estimate of the impulse response is then given by:
\begin{equation}
\label{eq: general bayes estimator}
\begin{split}
\hat{h}=\E[h|Y] &= \int h \, p(h|Y) \, dh\\
&= \int \int h \, p(h|\eta,Y) p(\eta|Y) \, dh d\eta\\
&= \int \E[h|Y,\eta] p(\eta|Y)\, d\eta
\end{split}
\end{equation}
where 
$ \E[h|Y,\eta]$
is the conditional estimate of $h$ when $\eta$ are fixed.
In a general framework these integrals are not analytically tractable and it is necessary to resort to effective approximations, e.g. analytical approximations or Markov Chain Monte Carlo (MCMC) methods. These  approximations yield to different approaches, such the so-called Empirical Bayes (EB) and Full Bayes (FB) estimators.
\begin{rem}
In principle, the estimator \eqref{eq: general bayes estimator} belongs to an infinite-dimensional space. However, for computational reasons, it is general practice to estimate a finite-length impulse response, whose length $n$ is chosen large enough to capture the dynamics of the estimated system. In this case, $h\in\mathbb{R}^n$ is modelled as a zero-mean Gaussian random vector with covariance $\bar{K}_\eta\in\mathbb{R}^{n\times n}$.
\end{rem}

\subsubsection{Empirical Bayes}
\label{subsec: EB Impulse response}
The Empirical Bayes approach is based on the assumption that the marginal posterior distribution on the hyperparameters $p(\eta|Y)$ can be approximated by a delta-function centered at its mode  $\hat\eta$; under this approximation  the outer integral in \eqref{eq: general bayes estimator} is trivially equal to $\E[h|Y,\eta] $ evaluated at $\hat\eta$. In order to estimate this value of $\eta$ the common approach is to consider a non informative prior on the hyperparameters and maximize the so-called marginal likelihood, $p(Y|\eta)$. Under the assumptions on the output noise and on the processes $\left\{y(t)\right\},\ \left\{u(t)\right\}$ (see  Section \ref{sec: Problem Formulation}), this marginal density can be computed in closed form, as discussed in \cite{SS2010} and \cite{GP-AC-GdN:11}, and is given by 
\begin{equation}
\label{eq: marginal log likelihood}
p(Y|\eta)= {\rm exp}\left( -\frac{1}{2} \ln(\det[2\pi\Sigma_y(\eta))- \frac{1}{2}Y^T\Sigma_y(\eta)^{-1}Y\right)
\end{equation}
\begin{equation}\label{Sigma_eta}
\Sigma_y(\eta) = \Phi \bar{K}_\eta \Phi^\top + \sigma^2I 
\end{equation}
where $\sigma^2:=Var\{e(t)\}$ is the  variance of the innovation process \eqref{eq: OE model} and $\Phi\in\mathbb{R}^{T\times n}$ is a matrix built with past input data; see \cite{ChenOL12}, \cite{GP-AC-GdN:11} for details.\\
It follows that we can compute the point  estimate of the hyperparameters for the EB approach as
\begin{equation}
\label{eq: eta ML estimator}
\hat{\eta}_{EB} = \arg \max_\eta p(Y|\eta)
\end{equation} 
and we can finally obtain the EB estimator of $h$, $\hat{h}_{EB} = \E [h|Y,{\hat \eta}_{EB}]$. Notice that, since $h(t)$ and $e(t)$ are Gaussian and independent, the convolution is a linear operation, then $Y$ and $h(t)$ are jointly Gaussian yielding also $h$ conditioned on $Y$ be Gaussian for a fixed $\eta$:
\begin{equation}\label{cond_posterior}
p(h|Y,\eta)\sim \mathcal{N}(\mu_h^{post}(\eta),\Sigma_h^{post}(\eta))
\end{equation}
where
\begin{eqnarray}
\mu_h^{post}(\eta)&=&\E [h|Y,\eta] \nonumber\\
&=& \bar{K}_{\eta}\Phi^\top \left(\Phi \bar{K}_{\eta} \Phi^\top + \sigma^2 I\right)^{-1}Y\label{eq: posterior mean}\\
\Sigma_h^{post}(\eta) &=& \bar{K}_{\eta} - \bar{K}_{\eta} \Phi^\top \Sigma_y(\eta)^{-1} \Phi \bar{K}_{\eta}\label{eq: posterior cov}
\end{eqnarray}
Hence the posterior estimate $\hat{h}_{EB}$ can be computed in closed form using \eqref{eq: posterior mean}.

\subsubsection{Full Bayes}
The Full Bayes approach has the advantage that it does not assume any particular distribution form of the marginal posterior $p(\eta|Y)$; therefore in principle, it generates a more accurate estimate than the EB estimator (under the assumption that the a priori Bayesian model is correct). As a disadvantage, in general it requires a much higher computational effort which,  when  the marginal posterior $p(\eta|Y)$ is sufficiently peaked, may not be counterbalanced by a significant performance increase.\\
Here we consider a \emph{full Bayes} estimator of the impulse response $h$ obtained by an adaptive version of the Metropolis-Hastings algorithm (Adaptive Metropolis, AM hereafter); see \cite{WG-SR-DS:96}, \cite{HH-ES-JT:01}. 

Recall that the target is to compute the posterior distribution of the impulse response given the data which, as mentioned in Section \ref{subsec: Bayesian Imp Resp Est}, cannot be computed analytically. For this reason, we tackle the problem by approximating the posterior as
\begin{equation}\label{ApproxPosterior} 
p(h|Y) = \int_\eta p(h|Y,\eta) p(\eta|Y)\, d\eta \simeq \frac{1}{N} \sum_{i=1}^{N}p(h|Y,\eta^{(i)})
\end{equation}
where $p(h|Y,\eta^{(i)})$ is the posterior density \eqref{cond_posterior} when the hyperparameters are fixed equal to $\eta^{(i)}$.\\
In order to do this, we need to design an MCMC algorithm to draw samples $\eta^{(i)}$ from $p(\eta|Y)$. Observe that:
\begin{equation}
\label{eq: marginal posterior hyper-parameters}
p(\eta\vert Y) = \frac{p(Y|\eta)p(\eta)}{p(Y)} \propto  p(Y\vert\eta)
\end{equation}
where we have assumed that $p(\eta)$ is a non informative prior distribution. Thus, by using \eqref{eq: marginal log likelihood} we can evaluate $p(\eta\vert Y)$ apart from the normalization constant $p(Y)$.\\
As  mentioned earlier, we have exploited the AM algorithm proposed in \cite{HH-ES-JT:01} to obtain the samples $\eta^{(i)}$. The basic idea which distinguishes the AM algorithm from a regular Metropolis-Hasting is to update the proposal distribution exploiting the new knowledge which becomes available: at each iteration $i$, the AM algorithm adopts a Gaussian proposal distribution centered at the previous sample $\eta^{(i-1)}$ and with a covariance matrix which is adaptively updated based on the samples $\eta^{(1)}, ..., \eta^{(i-1)}$. The updating recursion formula for the covariance matrix given in \cite{HH-ES-JT:01} is:
\begin{equation}
\label{eq: cov update}
\mc H_{i+1}= \frac{i-1}{i}\mc H_{i}+\frac{s_d}{i}(i\bar{\eta}^{(i)}\bar{\eta}^{(i)^\top} + \eta^{(i)} \eta^{(i)^\top} + \epsilon I_d)
\end{equation}
where $\bar{\eta}_k$ is the mean after $k$ samples, $s_d$ is a regularization parameter, $I_d$ is the identity matrix of dimension $d$, which is the dimension of the hyperparamters, and $\epsilon>0$ is an arbitrarily small constant.
The value of the regularization parameter initially has been chosen to be $s_d = \frac{2.4^2}{d}$, a value which gives good mixing properties in the Metropolis chain under the assumption of Gaussian targets and proposal, as shown in \cite{GAG-RGO-GWR:96}, then it has been empirically adjusted in order to have an acceptance rate of the MCMC algorithm around the 30\%.


The algorithm we implemented in order to obtain the FB estimate $\hat{h}_{FB}$ is briefly outlined in the following.

\vspace{3mm}

\begin{alg}
\mbox{\vspace{1truecm}}\\\textit{Sample hyperparameters through an AM algorithm}
\begin{enumerate}
\item Initialize the proposal density $q_i(\cdot)$ for the AM algorithm: set $q_0(\cdot) = \mathcal{N}(\hat{\eta}_{EB},\mathcal{H}_0)$, with
$$\mathcal{H}_0 = -\left[\frac{d^2 \ln[p(Y|\hat{\eta}_{EB})p(\hat{\eta}_{EB})]}{d\eta d\eta^T}\right]^{-1}$$
\item For $i>0$ Iterate:
\begin{itemize}
\item Sample $\eta$ from $q_i(\cdot \vert \eta^{(i-1)}) \thicksim \mathcal{N}(\eta^{(i-1)},\mc H_{i}))$
\item Sample $u$ from a uniform distribution on $[0,1]$
\item Set 
$$\eta^{(i)} = 
\left\lbrace
\begin{array}{ll}
\eta & \mbox{ if } u \leq \frac{p(Y|\eta)p(\eta)}{p(Y|\eta^{(i-1)})p(\eta^{(i-1)})}\\
\eta^{(i-1)} & \mbox{ otherwise}
\end{array}
\right.$$
\item Compute $\mc H_{i+1}$ according to equation \eqref{eq: cov update}.
\end{itemize}
\item After a (sufficiently long) burn-in period, keep the last $N$ samples $\eta^{(i)}$ which are (approximately) samples from   $p(\eta\vert Y)$.
\end{enumerate}
\textit{Estimate the impulse response:}
\begin{enumerate}\addtocounter{enumi}{3} 
\item For $i = 1$ to $N$ do
\begin{itemize}
\item Compute $\mu_h^{post}(\eta^{(i)}), \, \Sigma_{h}^{post}(\eta^{(i)})$ as in \eqref{eq: posterior mean}, \eqref{eq: posterior cov}.
\item Sample $h^{(i)}$ from $\mathcal{N} (\mu_h^{post}(\eta^{(i)}),\Sigma_{h}^{post}(\eta^{(i)}))$
\end{itemize}
\item The samples $h^{(i)}$ obtained above are  samples from $p(h|Y)$. The Minimum Variance estimate of $h$ is finally computed as:
\begin{equation}
\label{eq: MCMC posterior mean estimate}
\hat{h}_{FB}  = \frac{1}{N}\sum_{i=1}^N h^{(i)}
\end{equation}
\end{enumerate}
\end{alg}



\subsection{Confidence Set}
\label{subsec: Bayesian Confidence Set}

Within the Bayesian framework, the confidence of the final estimator is described  by the posterior density $p(h|Y)$. Since the Empirical Bayes (EB) and the Full Bayes (FB) estimators lead to different approximations of $p(h|Y)$, they will also lead to different definitions of the confidence set, as will be illustrated in the following.

\subsubsection{Empirical Bayes}
When the Emprical Bayes approach is considered, the posterior $p(h|Y)$ is the Gaussian distribution defined in \eqref{cond_posterior} with $\eta$ fixed to $\hat{\eta}_{EB}$.
Hence, one can define the following ellipsoidal confidence region in $\mathbb{R}^n$, with $n$ being the length of the estimated impulse response, i.e. $\hat{h}_{EB}\in\mathbb{R}^n$:
\begin{equation}\label{equ:ell_eb}
\mathcal{E}^{EB}_\alpha = \left\{x\in \mathbb{R}^n: (x-\hat{h}_{EB})^\top \Sigma_{\hat{\eta}_{EB}}^{-1} (x-\hat{h}_{EB}) \leq \chi_{\alpha}^2(n)\right\}
\end{equation} 
For a fixed probability level $\alpha$, $\chi_{\alpha}^2(n)$ is the value for which $\mbox{Pr}(\chi^2(n)<\chi_{\alpha}^2(n))=\alpha$. $\mathcal{E}^{EB}_\alpha$ defines the region in which a sample from $p(h|Y)$ will end up with probability $\alpha$. Note, for future use, that this set corresponds also to the set of ``size'' (= probability) $\alpha$ which satisfies:

\begin{equation}\label{eq:confidence:density}
p(h_{\mathcal{E}}|Y) \geq p(h_{\mathcal{E}^c}|Y) \quad \forall \quad h_{\mathcal{E}} \in \mathcal{E}^{EB}(\alpha)  \quad  h_{\mathcal{E}^c} \notin  \mathcal{E}^{EB}(\alpha) 
\end{equation}
To have a confidence set comparable to the ones defined for the classical methods, we approximate the set \eqref{equ:ell_eb} by a  point distribution obtained by sampling the posterior distribution $p(h|Y,\hat\eta_{EB})$ and retaining only the samples which belong to \eqref{equ:ell_eb}, that is:
\begin{equation}
S^{EB}_{\alpha} = \left\{ h^{(i)}\in \mathbb{R}^n: h^{(i)}\in \mathcal{E}^{EB}_\alpha \right\}, 
\end{equation}

\subsubsection{Full Bayes}
The FB estimator we previously described exploits the sample approximation to the posterior distribution in \eqref{ApproxPosterior}. Due to the non-Gaussianity of this approximated distribution, we can not define an ellipsoidal confidence region. However, an appropriate $\alpha$-level confidence set is given by:
\begin{equation}
S^{FB}_{\alpha} = \left\{h^{(i)}\in \mathbb{R}^n: \frac{1}{N}\sum_{j=1}^{N}p(h^{(i)}|Y,\eta_j) \geq p^{FB}_{\alpha}\right\}, 
\end{equation}
where $p^{FB}_{\alpha}$ is the $(1-\alpha)$-percentile of the set $$\left\{\frac{1}{N}\sum_{j=1}^{N}p(h^{(i)}|Y,\eta_j),\ i=1,...,N \right\}$$ That is, $S^{FB}_{\alpha}$ contains the impulse response samples $h^{(i)}$ associated with the $\alpha$-fraction of the highest values of the approximated posterior \eqref{ApproxPosterior}.


\section{Simulations} \label{sec: Simulations}
The performance of the described system identification approaches, EB, FB and PEM, are evaluated by using a Monte Carlo study over 100 datasets. At each run a model such as \eqref{eq: OE model} is estimated together with a confidence set around the estimated impulse response. The performance of the estimators are compared both in terms of 
impulse response fit as well as  of the accuracy of the corresponding confidence set, determined as illustrated in Sections \ref{sec: Parametric Identification Methods}  and \ref{sec: Non-Parametric Identification Methods}.

\subsection{Data}
The data-bank of system and input-output data used in our experiments have been already used and introduced in \cite{TC-MA-LL-AC-GP:14}. In particular, we applied the identification techniques to the data sets ``D4'' which is briefly described in the following.\\
The data set consists of 30th order random SISO dicrete-time systems having all the poles inside a circle of radius 0.95. These systems were simulated with a  unit variance  band-limited Gaussian signal with normalized band $[0,0.8]$. A zero mean white Gaussian noise,  with  variance adjusted so that the Signal to Noise Ration (SNR) is always equal to 1, was then added to the output data. The number of input-output data pairs is 500.\\
In addition, we experimented the dataset ``S1D2'' introduced in \cite{ChenOL12}. The results were similar to the ones obtained on dataset ``D4'' and outlined in the following; therefore, we are not going to report them here.
\subsection{Estimators}
\label{subsec:estimators_simulations}
\subsubsection{PEM} In the simulations we performed, the chosen model class for the PEM methods is OE. Model selection has been performed through BIC criterion, since it generally outperforms AIC. We will denote this estimator as PEM+BIC.\\
Moreover, as a reference we also consider an oracle estimator, denoted by PEM+OR, which has the (unrealistic) knowledge of the impulse response of the true system, $h$: among the OE models with complexity ranging from 2 to 30, it selects the one which gives the best fit to $h$.\\
\subsubsection{EB, FB}
For the Bayesian estimators illustrated in Section \ref{sec: Non-Parametric Identification Methods}, the choice of the prior distribution on the impulse response to be estimated is a crucial point for the identification problem.\\
The experiments we present in this Section have been obtained adopting a zero-mean Gaussian prior with a covariance matrix (kernel) given by the so-called ``DC''-kernel:
\begin{equation}
\bar K_\eta^{DC}(k,j) = c \rho^{\vert k-j \vert} \lambda^{(k+j)/2}
\end{equation}
where $c \geq 0$, $0 \leq \lambda \leq 1$ and $\vert \rho \vert \leq 1$ are the hyperparameters which form the set $\eta = \{c, \rho, \lambda\}$. For further details on the meaning of these hyperparameters and on the properties they induce in the estimated impulse response we refer to \cite{ChenOL12}, where the DC kernel has been proposed.

The length $n$ of the estimated impulse responses has been set to 100.

\vspace{2mm}

For ease of notation, we will now use the apex $X$ to denote a generic estimator among the ones previously illustrated, that is, PEM+BIC, PEM+OR, EB and FB.

\subsection{Impulse Response Fit}
As a first comparison, we would like to evaluate the ability of the considered identification techniques on the reconstruction of the true impulse response. Thus, for each estimated system and for each estimator $X$ we compute the so-called impulse response fit:
\begin{equation}
\label{eq:fit_imp_resp}
\mathcal{F}^X(\hat h) = 100 \times \Big(1- \frac{\Vert h - \hat{h} \Vert_2}{\Vert h \Vert_2}\Big) 
\end{equation}
where $h, \, \hat{h}$ are the true and the estimated impulse responses of the considered system.

\begin{figure}[!h]
\begin{center}
\includegraphics[scale=0.45]{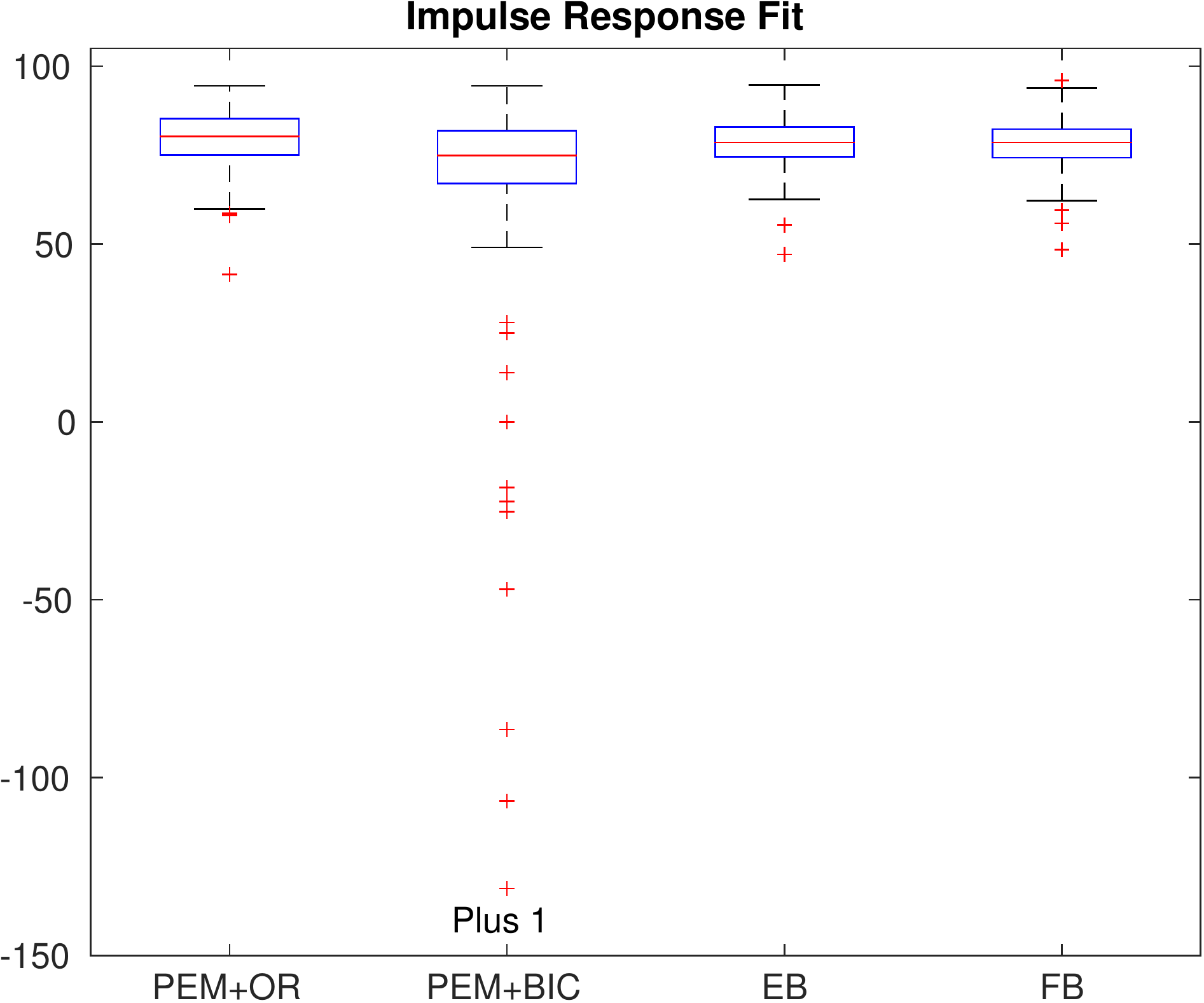} \caption{Monte Carlo results. Boxplots of the impulse response fit for the compared identification techniques.}
\label{fig: fit imp resp comparision}
\end{center}
\end{figure} 

Figure \ref{fig: fit imp resp comparision} displays the boxplots of index \eqref{eq:fit_imp_resp} for the 4 estimators and the resulting average can be seen in Tabel \ref{tab:fit_mean}.

\begin{table}[!h]
\begin{center}
\begin{tabular}{ccccc}
\toprule
& PEM+OR &  PEM+BIC & EB & FB\\
\midrule
Fit Mean & 79.1547 &  60.6676 &  78.4420 &  78.1332\\
\bottomrule
\end{tabular}
\caption{Comparison of average impulse response fit. }
\label{tab:fit_mean}
\end{center}
\end{table}

 The oracle estimator PEM+OR sets an upper bound on the achievable performance by a parametric methods;   we can note that EB performs remarkably well, with only a slightly inferior fit. The FB estimator performs similarly to EB, but it requires the implementation of a MCMC, which is highly computationally expensive. These results suggest that the marginal posterior $p(\eta\vert Y)$ is sufficiently well peaked to be approximated by a delta function (meaning that  $p(h \vert Y)\simeq p(h \vert Y,\hat\eta^{EB})$). The PEM+BIC estimator has weaker performances: a lower median and a long tail of systems with low fit are obtained. This is most likely due to the low pass characteristics of the input signal, which make the order estimation step particularly delicate. Indeed, in the dataset ``S1D2'' where the inputs were Gaussian white noises, PEM+BIC performed similar to the Bayesian estimators.
\subsection{Confidence Set Indexes}
The confidence sets which have been introduced in Sections \ref{subsec:pem_confidence_set} and \ref{subsec: Bayesian Confidence Set} are: $S^{PEM+OR+ASYMP}_\alpha$, $S^{PEM+OR+LIK}_\alpha$, $S^{PEM+BIC+ASYMP}_\alpha$, $S^{PEM+BIC+LIK}_\alpha$, $S^{EB}_\alpha$ and $S^{FB}_\alpha$. As before,$S^{X}_\alpha$ will generically denote one of them.\\
In the simulations we present, the previously defined confidence sets are made of $N=7200$ samples and we set $\alpha = 0.95$.  These are only approximations of 	a ``true'' $\alpha$-level confidence set and thus our aim is to study how well they perform both in term  of ``coverage'' (how often does the $\alpha$-level confidence set contain the ``true'' value?) as well as of size (how big is an $\alpha$-level confidence set?). Unfortunately, since our sets are only defined through a set of points, it is not possible to define a notion of inclusion (does the true system belong to the confidence set?) and as a proxy to this we thus consider the following index which measures the relative distance from the true system and the closest point within the confidence set: 

\begin{enumerate}
\item \textit{Coverage  Index}: For a fixed probability level $\alpha$, it is given by
\begin{equation}
\label{def:min_dist_point}
\mathcal{I}^X_1(\alpha) := \min_{x \in S^X_\alpha} \frac{\|x-h\|}{\|h\|}
\end{equation}
where $h$ denotes the true impulse response. For future analysis the usage of the concept ``coverage'' will be meant as in definition \eqref{def:min_dist_point}.

As far as the ``size'' of the confidence sets we consider the index:

\item \textit{Confidence Set Size}: It evaluates the area of the interval which includes the whole slot of impulse responses contained in $S^X_\alpha$. Let us define the vectors $\bar h^X \in \R^n$ and $\underline h^X \in \R^n$ whose $j$-entries are $\bar h^X(j) := \max_i h^{(i)}(j)$ and $\underline h^X(j) := \min_i h^{(i)}(j)$, respectively, with $h^{(i)} \in S^X_\alpha$; the index we consider is defined as:
\begin{equation}
\label{def:area}
\mathcal{I}^X_2(\alpha) = \sum_{j=1}^{n} \bar h^X(j) - \underline h^X(j)
\end{equation}

\begin{figure}[!h]
\begin{center}
\includegraphics[scale=0.45]{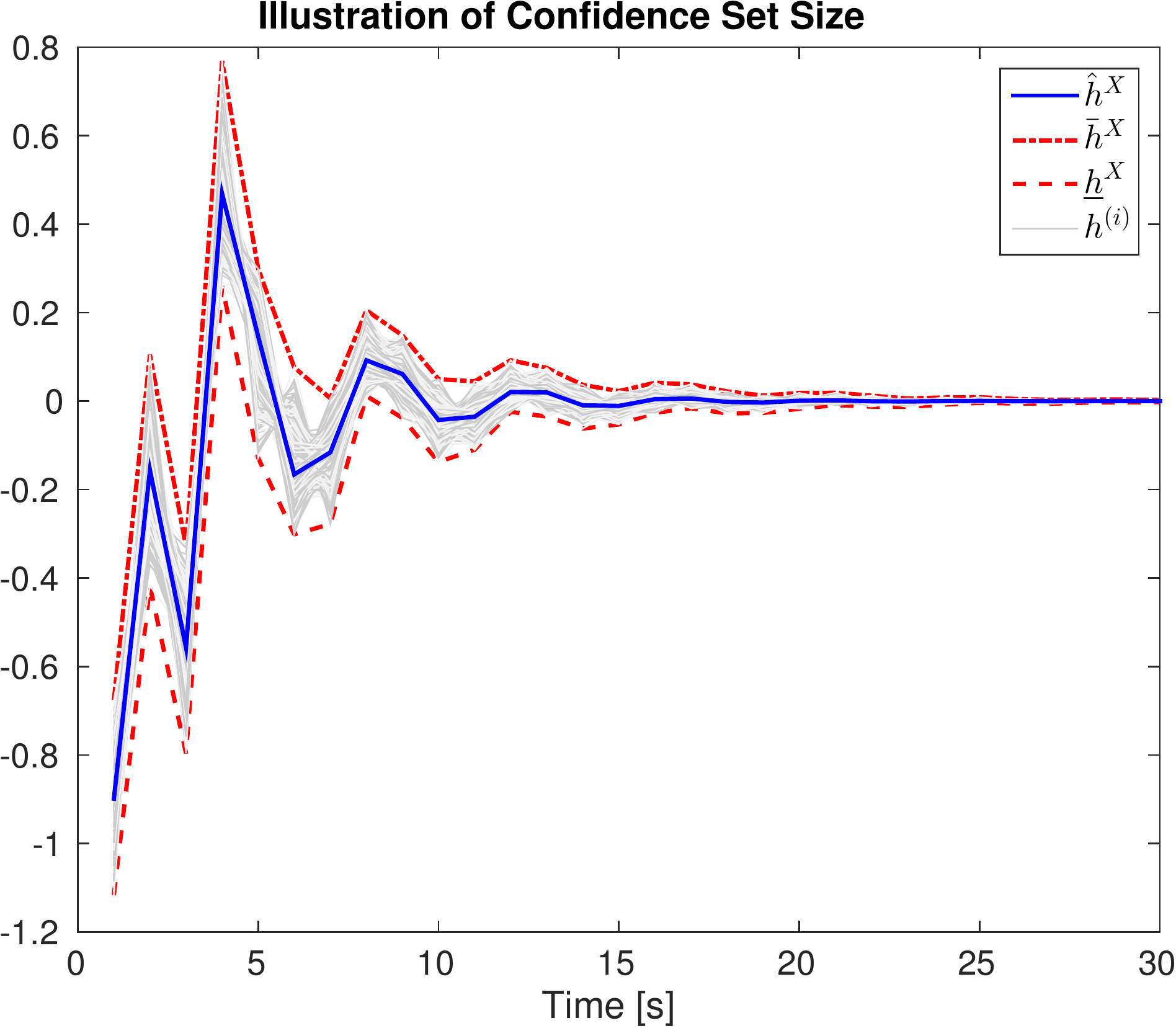} 
\caption{Illustration of the idea of the \textit{Confidence set size} index for a single system.}
\label{fig:area_single_sys}
\end{center}
\end{figure} 
\end{enumerate}

Referring to Figure \ref{fig:area_single_sys}, a large confidence set is more likely to  contain the true impulse response, giving a low value of $\mathcal{I}^X_1(\alpha)$, but it will also denote a large amount of uncertainty in the returned estimate, thus leading to a large value of $\mathcal{I}^X_2(\alpha)$.

\begin{figure}[!h]
\begin{center}
\includegraphics[scale=0.45]{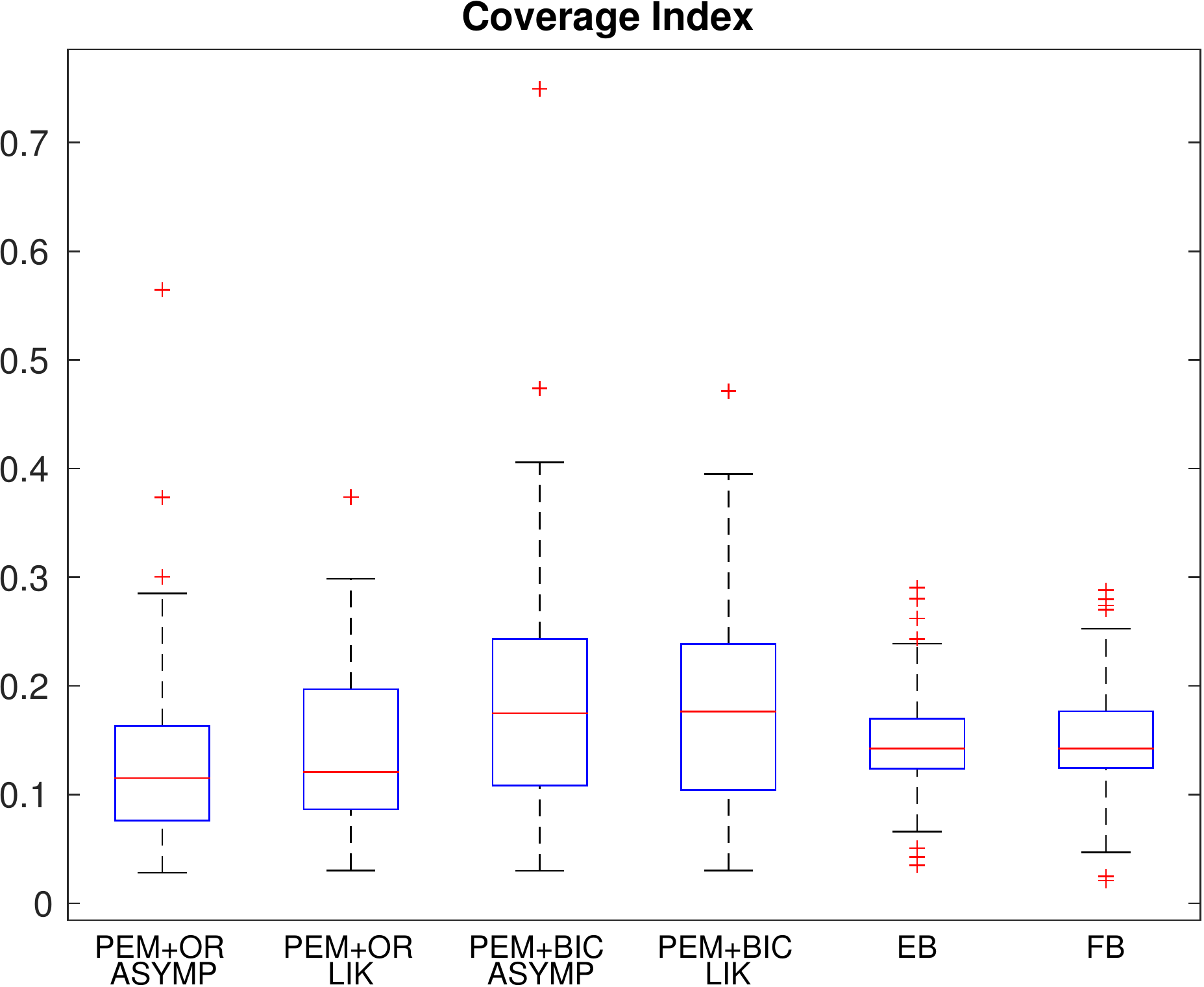} 
\caption{Monte Carlo results. Boxplots of the \textit{Coverage Index} for the compared identification techniques.}
\label{fig:min_dist_boxplot}
\end{center}
\end{figure} 

Figure \ref{fig:min_dist_boxplot} illustrates the boxplots for index \eqref{def:min_dist_point}.
 {The confidence sets of the oracle perform well in terms of coverage,  which is rather obvious because the estimator is selected by the oracle if its relative distance to the true system is small.  EB and FB provide very similar performance in terms of coverage, outperforming   the confidence sets computed from the parametric approach endowed with  BIC.


\begin{figure}[!h]
\begin{center}
\includegraphics[scale=0.45]{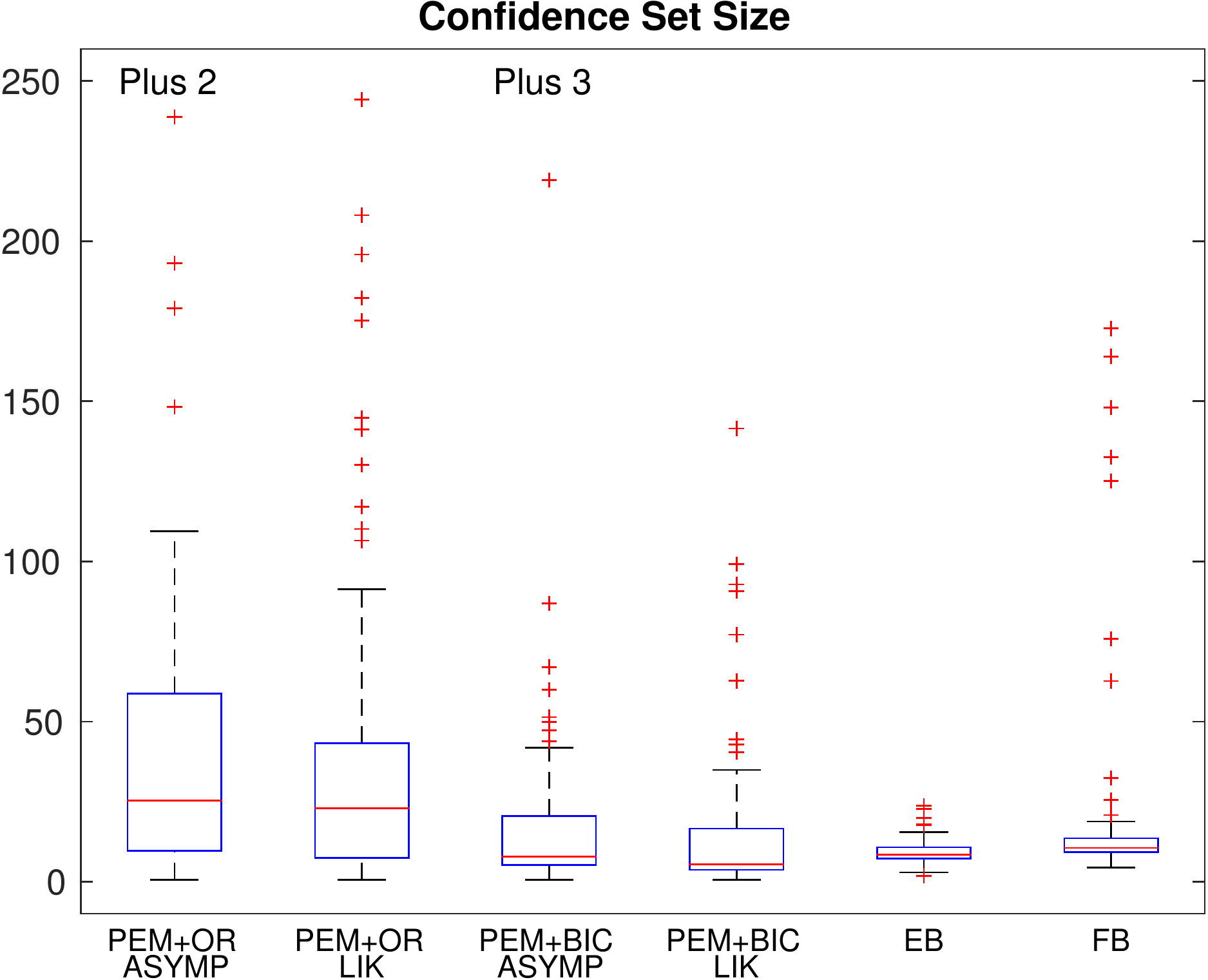} 
\caption{Monte Carlo results. Boxplots of the \textit{Confidence Set Size} for the compared identification techniques.}
\label{fig:area_boxplot}
\end{center}
\end{figure} 

Figure \ref{fig:area_boxplot} illustrates the boxplots for index \eqref{def:area}. The EB confidence set has a remarkably smaller size than the others. The size of the FB 
confidence set is slightly larger  than the EB one, which is rather obvious since also uncertainty related to the hyperparameters is accounted for. 
The parametric methods PEM+OR and PEM+BIC have larger confidence sets than the Bayesian ones. In particular, the two PEM+OR confidence sets are larger than the ones returned by the PEM+BIC estimator: this can be explained from the fact that PEM+OR tends to select higher-order models, thus bringing more uncertainty into the estimated systems. Comparing the \textit{Asymptotic} and the \textit{Likelihood Sampling} confidence sets it is clear that  the latter is slightly more precise than the former. This is due to the fact that the \textit{Asymptotic} confidence set is an approximation which holds for large data sets, while the \textit{Likelihood Sampling} is correct for any finite sample size; however, this improvement comes at a rather high computational price needed to run the MCMC  sampler. It is important to note that the asymptotic theory does not take into account stability issues: namely, the confidence set derived from the Gaussian asymptotic distribution \eqref{eq:asdistr} could contain unstable impulse responses. Therefore the sampling procedure described in Section \ref{subsubsec:asymp_pem_confset} could yield to diverging confidence set size. In order to avoid this problem we truncated the asymptotic Gaussian distribution within the stability region. Clearly, this fact shows an intrinsic problem of the asymptotic theory. We should also like to recall that the asymptotic as well as likelihood based confidence intervals do not account for uncertainty in the order estimation step.\\

By comparing the results in both Figures \ref{fig:min_dist_boxplot}-\ref{fig:area_boxplot} we can conclude that: among the feasible identification methods, EB and FB are preferable both in terms of coverage as well as size. In this case there seems to be no gain in using the much more computationally expensive FB. It seems also fair to say that the confidence sets attached to the parametric approaches, even those of the oracle estimators, are significantly worse than those obtained from the Bayesian methods. 
Not surprisingly, focusing on the parametric  confidence sets, the ones obtained through sampling techniques are significantly smaller than the ``asymptotic'' counterparts, but  at the cost of an extra MCMC algorithm. 


\begin{rem}
At this point one could argue that the sets $S^X_\alpha$ are only ``sample'' approximations of a confidence set, while one may be interested in having a bounded region as a confidence set. In the case of the EB estimator this region is directly defined since the    posterior distribution is Gaussian, thus naturally leading to ellipsoidal confidence regions \eqref{equ:ell_eb}. For all the other estimators, it is in principle possible to build outer approximations of the confidence sets e.g. building a  minimum size set which includes all the points in $S^X_\alpha$; examples are the  convex hull or an ellipsoid.\\
The convex hull can be computed with off-the-shelf algorithms (such as the  Matlab routine \verb!convhulln.m!), while the smallest ellipsoid (in terms of sum of squared semi-axes length) can be found solving the following problem:
\begin{eqnarray}\label{equ:fit_ellips}
P_{\alpha}^{opt},c_{\alpha}^{opt} :=  &\arg & \min_{P,c}  \mbox{Trace}\ P\nonumber\\
&s.t.& \left[\begin{array}{cc} P & (h^{(i)}-c)\\ (h^{(i)}-c)^\top & 1 \end{array}\right] \succ 0,\nonumber\\
& h^{(i)}&\in S^{X}_{\alpha}
\end{eqnarray}
See \cite{GC:02} for further details. The corresponding ellipsoid is then given by
\begin{equation}
\mathcal{E}^{opt}_\alpha = \left\{x \in \mathbb{R}^n: (x-c_{\alpha}^{opt})^\top (P_{\alpha}^{opt})^{-1} (x-c_{\alpha}^{opt}) \leq 1  \right\}
\end{equation}
However, the computation of the convex hull as well as the solution of the optimization problem \eqref{equ:fit_ellips}  become computationally intractable for moderate ambient space and sample sizes. E.g.  when the impulse response lives in $\mathbb{R}^n$, $n=100$, the set   $S^{X}_{\alpha}$ contains $N=7200$  this computations are prohibitive with off-the-shelf methods. To overcome this issue, we tried to approximate the optimal ellipsoid $\mathcal{E}^{opt}_\alpha$ by using the sample mean $\bar{h}_{S^{X}_{\alpha}}$ and the sample covariance $\Sigma_{S^{X}_{\alpha}}$ of the elements in $S^{X}_{\alpha}$; namely:
\begin{eqnarray}\label{equ:ellips_approx}
\mathcal{E}^{X}_\alpha &=&
 \left\{x \in \mathbb{R}^n: (d_\alpha^{X} )^\top \Sigma_{S^{X}_{\alpha}}^{-1} d_\alpha^{X} \leq k^{X}_{\alpha}  \right\},\nonumber\\
 d_\alpha^{X} &=& x-\bar{h}_{S^{X}_{\alpha}}
\end{eqnarray}
where $k_{\alpha}^{X}$ is a constant appropriately chosen so that all the elements of $S^{X}_{\alpha}$ fall within $\mathcal{E}_{\alpha}^{X}$. However, it can be observed that these ellipsoids are rather rough approximations of the sets $S^{X}_{\alpha}$. E.g., inspecting 2D sections of the $n$-dimensional ellipsoids, it can be seen that often the axis orientation was not correct, thus leading to sets which are much larger than needed.  This fact was mainly observed for the confidence sets related to PEM estimates.\\
These observations suggest  that the quality of the confidence sets obtained through the ellipsoidal approximation \eqref{equ:ellips_approx} would have been highly dependent on the quality of the fitted ellipsoid. Therefore, we concluded that a comparison among the different estimators, based on this kind of confidence set, would have led to unreliable results; therefore such results have not been reported. 

\end{rem}


\section{Conclusions} \label{sec: Conclusions}

We have presented an in-depth comparison between parametric and Bayesian methods for system identification. Our results complement previous  findings showing that Bayesian methods not only outperform parametric methods in terms of point estimators, but also provide better  approximations for uncertainty regions. From our  limited experience there seems to be very little advantage in using Full Bayes approaches which entail a much higher computational load than Empirical Bayes methods. It is interesting to note that Bayesian estimators and their confidence sets are competitive even with the parametric methods equipped with an oracle which has the knowledge of the true impulse response. In addition, with regard to the parametric techniques, we showed that the confidence sets obtained through sampling techniques improve the ones returned by the ``asymptotic'' approximation.

%


\bibliographystyle{IEEEtran}
\bibliography{References}

\end{document}